\documentclass[10pt,twocolumn,letterpaper]{article}

\usepackage{cvpr}  

\usepackage{bbm}
\usepackage{graphicx}
\usepackage{amsmath}
\usepackage{amssymb}
\usepackage{amsthm}
\usepackage{mathrsfs}
\usepackage{amsfonts}
\usepackage{booktabs}
\usepackage{multirow}
\usepackage{multicol}
\usepackage[table]{xcolor}
\usepackage{color, colortbl}
\usepackage{pifont}
\newcommand{\cmark}{\ding{51}}%

\usepackage[pagebackref,breaklinks,colorlinks]{hyperref}

\usepackage[capitalize]{cleveref}
\crefname{section}{Sec.}{Secs.}
\Crefname{section}{Section}{Sections}
\Crefname{table}{Table}{Tables}
\crefname{table}{Tab.}{Tabs.}

\def\eg{\emph{e.g.}} 
\def\ie{\emph{i.e.}} 
\def\vs{\emph{vs.}}
\def\etc{\emph{etc.}}

\begin{document}

\title{VAD: \underline{V}ectorized Scene Representation for Efficient \underline{A}utonomous \underline{D}riving}
\author{Bo Jiang$^{1,\star,\diamond}$,\quad Shaoyu Chen$^{1,\star,\diamond}$,\quad Qing Xu$^2$,\quad Bencheng Liao$^{1,\diamond}$, \quad Jiajie Chen$^2$,\\
\quad Helong Zhou$^2$,\quad Qian Zhang$^2$,\quad Wenyu Liu$^1$, \quad Chang Huang$^2$, \quad Xinggang Wang$^{1,\boxtimes}$\\	
[2mm]
$^1$~\normalsize{Huazhong University of Science \& Technology} \quad	
$^2$~\normalsize{Horizon Robotics}\\	
{\tt\small \{bjiang, shaoyuchen, bcliao, liuwy, xgwang\}@hust.edu.cn}\\	
{\tt\small \{qing.xu, jiajie.chen, helong.zhou, qian01.zhang, chang.huang\}@horizon.ai}\\
\normalsize{
\url{https://github.com/hustvl/VAD}
}
}

\maketitle

\let\thefootnote\relax\footnotetext{$^\star$ Equal contribution; $^\diamond$ Interns of Horizon Robotics when doing this work; $^\boxtimes$ Corresponding author: \texttt{xgwang@hust.edu.cn}}

\begin{abstract}
Autonomous driving requires a comprehensive understanding of the surrounding environment for reliable trajectory planning. Previous works rely on dense rasterized scene representation (\eg, agent occupancy and semantic map) to perform planning, which is computationally intensive and misses the instance-level structure information. In this paper, we propose VAD, an end-to-end vectorized paradigm for autonomous driving, which models the driving scene as a fully vectorized representation. The proposed vectorized paradigm has two significant advantages. On one hand, VAD exploits the vectorized agent motion and map elements as explicit instance-level planning constraints which effectively improves planning safety. On the other hand, VAD runs much faster than previous end-to-end planning methods by getting rid of computation-intensive rasterized representation and hand-designed post-processing steps. VAD achieves state-of-the-art end-to-end planning performance on the nuScenes dataset, outperforming the previous best method by a large margin. Our base model, VAD-Base, greatly reduces the average collision rate by 29.0\% and runs 2.5$\times$ faster. Besides, a lightweight variant, VAD-Tiny, greatly improves the inference speed (up to 9.3$\times$) while achieving comparable planning performance. We believe the excellent performance and the high efficiency of VAD are critical for the real-world deployment of an autonomous driving system. Code and models are available at \url{https://github.com/hustvl/VAD} for facilitating future research. 
\end{abstract}

\section{Introduction}
\label{sec:intro}
Autonomous driving requires both comprehensive scene understanding for ensuring safety and high efficiency for real-world deployment. An autonomous vehicle needs to efficiently perceive the driving scene and perform reasonable planning based on the scene information.

Traditional autonomous driving methods~\cite{kendall2019learning, gonzalez2015review, chen2021data, xu2021autonomous} adopt a modular paradigm, where perception and planning are decoupled into standalone modules. The disadvantage is, the planning module cannot access the original sensor data, which contains rich semantic information. And since planning is fully based on preceding perception results, the error in perception may severely influence planning and can not be recognized and cured in the planning stage, which leads to the safety problem. 

\begin{figure}[]
    \centering
    \includegraphics[width=\linewidth]{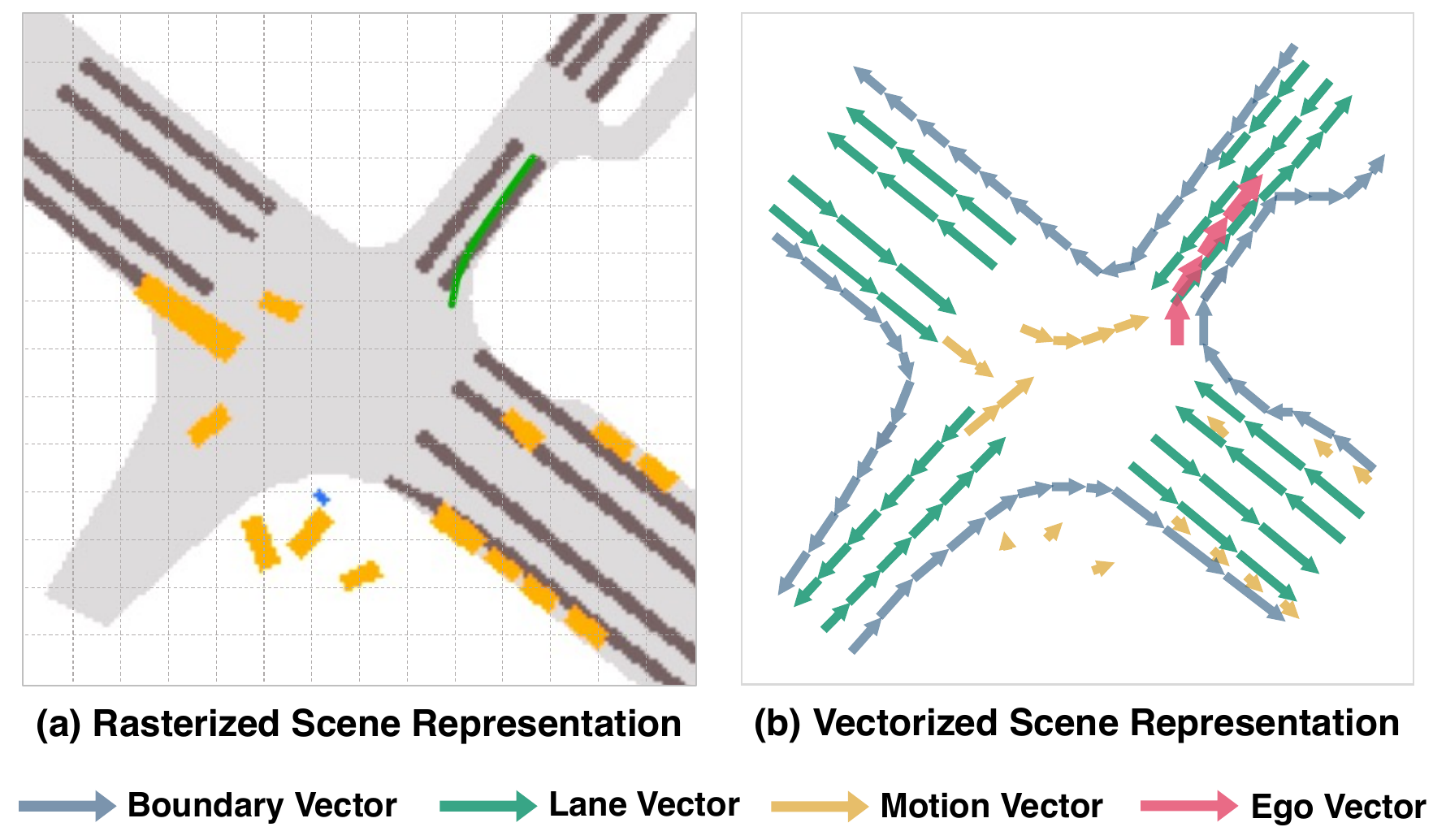}
    \caption{Previous paradigms mainly rely on rasterized representation (a) for planning (\eg, semantic map, occupancy map, flow map, and cost map), which is computationally intensive. The proposed VAD is fully based on the vectorized scene representation (b) for end-to-end planning. VAD leverages instance-level structure information as planning constraints and guidance, achieving promising performance and efficiency.}
    \label{fig:intro}
\end{figure}

Recently, end-to-end autonomous driving methods ~\cite{hu2022stp3, hu2022uniad, casas2021mp3, cui2021lookout} take sensor data as input for perception and output planning results with one holistic model. Some works~\cite{pomerleau1988alvinn, codevilla2019exploring, prakash2021multi} directly output planning results based on the sensor data without learning scene representation, which lacks interpretability and is difficult to optimize. Most works~\cite{hu2022stp3, hu2022uniad, casas2021mp3} transform the sensor data into rasterized scene representation (\eg, semantic map, occupancy map, flow map, and cost map) for planning.  
Though straightforward, rasterized representation is computationally intensive and misses critical instance-level structure information.

In this work, we propose VAD (\textbf{V}ectorized \textbf{A}utonomous \textbf{D}riving), an end-to-end vectorized paradigm for autonomous driving. VAD models the scene in a fully vectorized way (\ie, vectorized agent motion and map), getting rid of computationally intensive rasterized representation.

We argue that vectorized scene representation is superior to rasterized one. Vectorized map (represented as boundary vectors and lane vectors) provides road structure information (\eg, traffic flow, drivable boundary, and lane direction), and helps the autonomous vehicle narrow down the trajectory search space and plan a reasonable future trajectory. The motion of traffic participants (represented as agent motion vectors) provides instance-level restriction for collision avoidance. What's more, vectorized scene representation is efficient in terms of computation, which is important for real-world applications.

VAD takes full advantage of the vectorized information to guide planning both implicitly and explicitly. On one hand,
VAD adopts map queries and agent queries to implicitly learn instance-level map features and agent motion features from sensor data, and extracts guidance information for planning via query interaction. On the other hand, VAD proposes three instance-level planning constraints based on the explicit vectorized scene representation: the ego-agent collision constraint for maintaining a safe distance between the ego vehicle and other dynamic agents both laterally and longitudinally; the ego-boundary overstepping constraint for pushing the planning trajectory away from the road boundary; and the ego-lane direction constraint for regularizing the future motion direction of the autonomous vehicle with vectorized lane direction. Our proposed framework and the vectorized planning constraints effectively improve the planning performance, without incurring large computational overhead.

Without fancy tricks or hand-designed post-processing steps, VAD achieves state-of-the-art (SOTA) end-to-end planning performance and the best efficiency on the challenging nuScenes~\cite{caesar2020nuscenes} dataset. Compared with the previous SOTA method UniAD~\cite{hu2022uniad}, our base model, VAD-Base, greatly reduces the average planning displacement error by 30.1\% (1.03m \vs\ 0.72m) and the average collision rate by 29.0\% (0.31\% \vs\ 0.22\%), while running 2.5$\times$ faster (1.8 FPS \vs\ 4.5 FPS). The lightweight variant, VAD-Tiny, runs 9.3$\times$ faster (1.8 FPS \vs\ 16.8 FPS) while achieving comparable planning performance, the average planning displacement error is 0.78m and the average collision rate is 0.38\%. We also demonstrate the effectiveness of our design choices through thorough ablations.

Our key contributions are summarized as follows:
\begin{itemize}
\item We propose VAD, an end-to-end vectorized paradigm for autonomous driving. VAD models the driving scene as a fully vectorized representation, getting rid of computationally intensive dense rasterized representation and hand-designed post-processing steps.
\item VAD implicitly and explicitly utilizes the vectorized scene information to improve planning safety, via query interaction and vectorized planning constraints.
\item VAD achieves SOTA end-to-end planning performance, outperforming previous methods by a large margin. Not only that, because of the vectorized scene representation and our concise model design, VAD greatly improves the inference speed, which is critical for the real-world deployment of an autonomous driving system.
\end{itemize}

It's our belief that autonomous driving can be performed in a fully vectorized manner with high efficiency. We hope the impressive performance of VAD can reveal the potential of vectorized paradigm to the community.

\section{Related Work}
\label{sec:related}

\paragraph{Perception.} Accurate perception of the driving scene is the basis for autonomous driving. We mainly introduce some camera-based 3D object detection and online mapping methods which are most relevant to this paper. DETR3D~\cite{wang2022detr3d} uses 3D queries to sample corresponding image features and accomplish detection without non-maximum suppression. PETR~\cite{liu2022petr} introduces 3D positional encoding to image features and uses detection queries to learn object features via attention~\cite{vaswani2017attention} mechanism. Recently, bird's-eye view (BEV) representation has become popular and has greatly contributed to the field of perception~\cite{li2022bevformer, zhang2022beverse, hu2021fiery, liao2022maptr,gkt,polarbev,lanegap}. LSS~\cite{philion2020lss} is a pioneering work that introduces depth prediction to project features from perspective view to BEV. BEVFormer~\cite{li2022bevformer} proposes spatial and temporal attention for better encoding the BEV feature map and achieves remarkable detection performance with camera input only. FIERY~\cite{hu2021fiery} and BEVerse~\cite{zhang2022beverse} use the BEV feature map to predict dense map segmentation. HDMapNet~\cite{li2022hdmapnet} converts lane segmentation to vectorized map with post-processing steps. VectorMapNet~\cite{liu2022vectormapnet} predicts map elements in an autoregressive way. MapTR~\cite{liao2022maptr} recognizes the permutation invariance of the map instance points and can predict all map elements simultaneously. 
LaneGAP~\cite{lanegap} models the lane graph in a novel path-wise manner, which well preserves the continuity of the lane and encodes traffic information for planning.
We leverage a group of BEV queries, agent queries, and map queries to accomplish scene perception following BEVFormer~\cite{li2022bevformer} and MapTR~\cite{liao2022maptr}, and further use these query features and perception results in the motion prediction and planning stage. Details are shown in Sec.~\ref{sec:method}.

\begin{figure*}[h]
\centering
\includegraphics[width=0.98\textwidth]{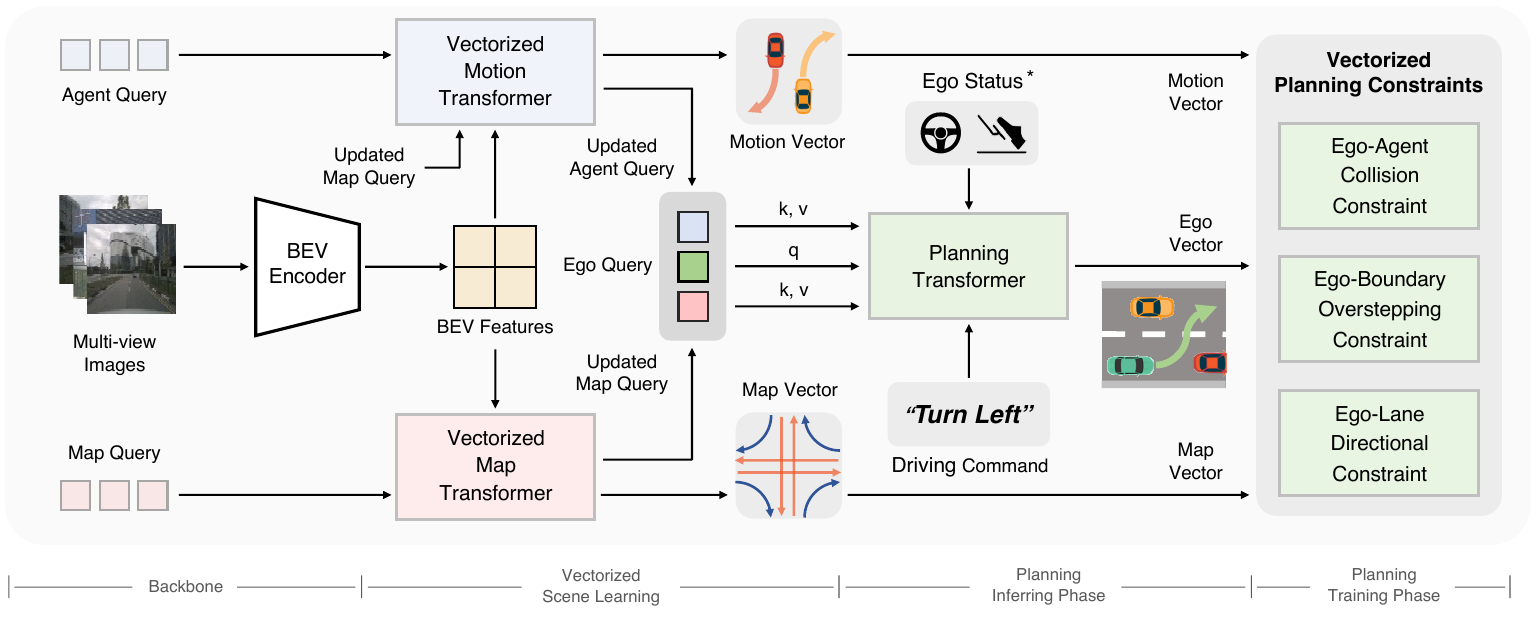} 
\vspace{2mm}
\caption{\textbf{Overall architecture of VAD.} The full pipeline of VAD is divided into four phases. Backbone includes an image feature extractor and a BEV encoder to project the image features to the BEV features. Vectorized Scene Learning aims to encode the scene information into agent queries and map queries, as well as represent the scene with motion vectors and map vectors. In the inferring phase of planning, VAD utilizes an ego query to extract map and agent information via query interaction and outputs the planning trajectory (represented as ego vector). The proposed vectorized planning constraints regularize the planning trajectory in the training phase. *: optional.}
\label{fig:arch}
\end{figure*}

\paragraph{Motion Prediction.} Traditional motion prediction takes perception ground truth (\eg, agent history trajectories and HD map) as input. Some works~\cite{chai2019multipath, phan2020covernet} render the driving scene as BEV images and adopt CNN-based networks to predict future motion. Some other works~\cite{gao2020vectornet, liu2021mmtrans, ngiam2021scenetrans} use vectorized representation, and adopt GNN~\cite{liang2020lanegcn} or Transformer~\cite{vaswani2017attention, liu2021mmtrans, ngiam2021scenetrans} to accomplish learning and prediction. Recent end-to-end works~\cite{hu2021fiery, zhang2022beverse, gu2022vip3d, jiang2022pip} jointly perform perception and motion prediction. Some works~\cite{hu2021fiery, zhang2022beverse, hu2022hope} see future motion as dense occupancy and flow instead of agent-level future waypoints. ViP3D~\cite{gu2022vip3d} predicts future motion based on the tracking results and HD map. PIP~\cite{jiang2022pip} proposes an interaction scheme between dynamic agents and static vectorized map, and achieves SOTA performance without relying on HD map. VAD learns vectorized agent motion by interacting between dynamic agents and static map elements, inspired by~\cite{jiang2022pip}.

\paragraph{Planning.}  Recently, learning-based planning methods prevail. Some  works~\cite{pomerleau1988alvinn, codevilla2019exploring, prakash2021multi} omit intermediate stages such as perception and motion prediction, and directly predict planning trajectories or control signals. Although this idea is straightforward and simple, they lack interpretability and are difficult to optimize. Reinforcement learning is quite up to the planning task and has become a promising research direction~\cite{toromanoff2020end, chen2021learning, chekroun2021gri}. Explicit dense cost map has great interpretability and is widely used~\cite{casas2021mp3, hu2022stp3, cui2021lookout, sadat2020p3}. The cost maps are constructed from the perception or motion prediction results or come from a learning-based module. And hand-crafted rules are often adopted to select the best planning trajectory with minimum cost. The construction of a dense cost map is computationally intensive and the using of hand-crafted rules brings robustness and generalization problems. UniAD~\cite{hu2022uniad} effectively incorporates the information provided by various preceding tasks to assist planning in a goal-oriented spirit, and achieves remarkable performance in perception, prediction, and planning. PlanT~\cite{renz2022plant} takes perception ground truth as input and encodes the scene in object-level representation for planning. In this paper, we explore the potential of vectorized scene representation for planning and get rid of dense maps or hand-designed post-processing steps.

\section{Method}
\label{sec:method}
\paragraph{Overview.}
The overall framework of VAD is depicted in Fig.~\ref{fig:arch}. Given multi-frame and multi-view image input, VAD first encodes the image features with a backbone network and utilizes a group of BEV queries to project the image features to the BEV features~\cite{li2022bevformer, philion2020lss,zhou2022cvt}. 
Second, VAD utilizes a group of agent queries and map queries to learn the vectorized scene representation, including vectorized map and vectorized agent motion (Sec.~\ref{sec:representation}). Third, planning is performed based on the scene information (Sec.~\ref{sec:planning}). 
Specifically, VAD uses an ego query to learn the implicit scene information through interaction with agent queries and map queries. Based on the ego query, ego status features, and high-level driving command, the Planning Head outputs the planning trajectory. Besides, VAD introduces three vectorized planning constraints to restrict the planning trajectory at the instance level (Sec.~\ref{sec:plan_loss}). VAD is fully differentiable and trained in an end-to-end manner (Sec.~\ref{sec:end-to-end-learning}).

\subsection{Vectorized Scene Learning}
Perceiving traffic agents and map elements are important in driving scene understanding. VAD encodes the scene information into query features and represents the scene by map vectors and agent motion vectors.
\label{sec:representation}

\paragraph{Vectorized Map.}  Previous works~\cite{hu2022stp3, hu2022uniad} use rasterized semantic maps to guide the planning, which misses critical instance-level structure information of the map. VAD utilizes a group of map queries~\cite{liao2022maptr} $Q_{m}$ to extract map information from BEV feature map and predicts map vectors $\hat{V}_m \in \mathbb{R}^{N_m \times N_p \times 2}$ and the class score for each map vector, where $N_m$ and $N_p$ denote the number of predicted map vectors and the points contained in each map vector. Three kinds of map elements are considered: lane divider, road boundary, and pedestrian crossing. The Lane divider provides direction information, and the road boundary indicates the drivable area. Map queries and map vectors are both leveraged to improve the planning performance (Sec.~\ref{sec:planning} and Sec.~\ref{sec:plan_loss}).

\paragraph{Vectorized Agent Motion.}
VAD first adopts a group of agent queries $Q_{a}$ to learn agent-level features from the shared BEV feature map via deformable attention~\cite{zhu2020deformable}. The agent's attributes (location, class score, orientation, \etc) are decoded from the agent queries by an MLP-based decoder head. 
To enrich the agent features for motion prediction,
VAD performs agent-agent and agent-map interaction~\cite{jiang2022pip, ngiam2021scenetrans} via attention mechanism. Then VAD predicts future trajectories of each agent, represented as multi-modality motion vectors $\hat{V}_a \in \mathbb{R}^{N_a \times N_k \times T_f \times 2}$. $N_a, N_k$, and $T_f$ denote the number of predicted agents, the number of modalities, and the number of future timestamps. Each modality of the motion vector indicates a kind of driving intention. VAD outputs a probability score for each modality. The agent motion vectors are used to restrict the ego planning trajectory and avoid collision (Sec.~\ref{sec:plan_loss}). Meanwhile, the agent queries are sent into the planning module as scene information (Sec.~\ref{sec:planning}). 

\subsection{Planning via Interaction}
\label{sec:planning}

\paragraph{Ego-Agent Interaction.}
VAD utilizes a randomly initialized ego query $Q_{\rm ego}$ to learn the implicit scene features which are valuable for planning. In order to learn the location and motion information of other dynamic traffic participants, the ego query first interacts with the agent queries through a Transformer decoder~\cite{vaswani2017attention}, in which
ego query serves as query of attention $q$, and agent queries serve as key $k$ and value $v$. 
The ego position $p_{\rm ego}$ and agent positions $p_{a}$ predicted by the perception module are encoded by a single layer MLP ${\rm PE}_{1}$, and serve as query position embedding $q_{\rm pos}$ and key position embedding $k_{\rm pos}$. The positional embeddings provide information on the relative position relationship between agents and the ego vehicle. The above process can be formulated as:

\begin{equation}
\begin{aligned}
\label{eq:ego-agent}
    &Q_{\rm ego}^{'} = {\rm TransformerDecoder}(q, \, k, \, v, \, q_{\rm pos}, \, k_{\rm pos}), \\
    &q = Q_{\rm ego}, \, k = v = Q_{a}, \\
    &q_{\rm pos} = {\rm PE}_{1}(p_{\rm ego}), \, k_{\rm pos}={\rm PE}_{1}(p_{a}).
\end{aligned}
\end{equation}

\begin{figure*}[h]
\centering
\includegraphics[width=0.9\textwidth]{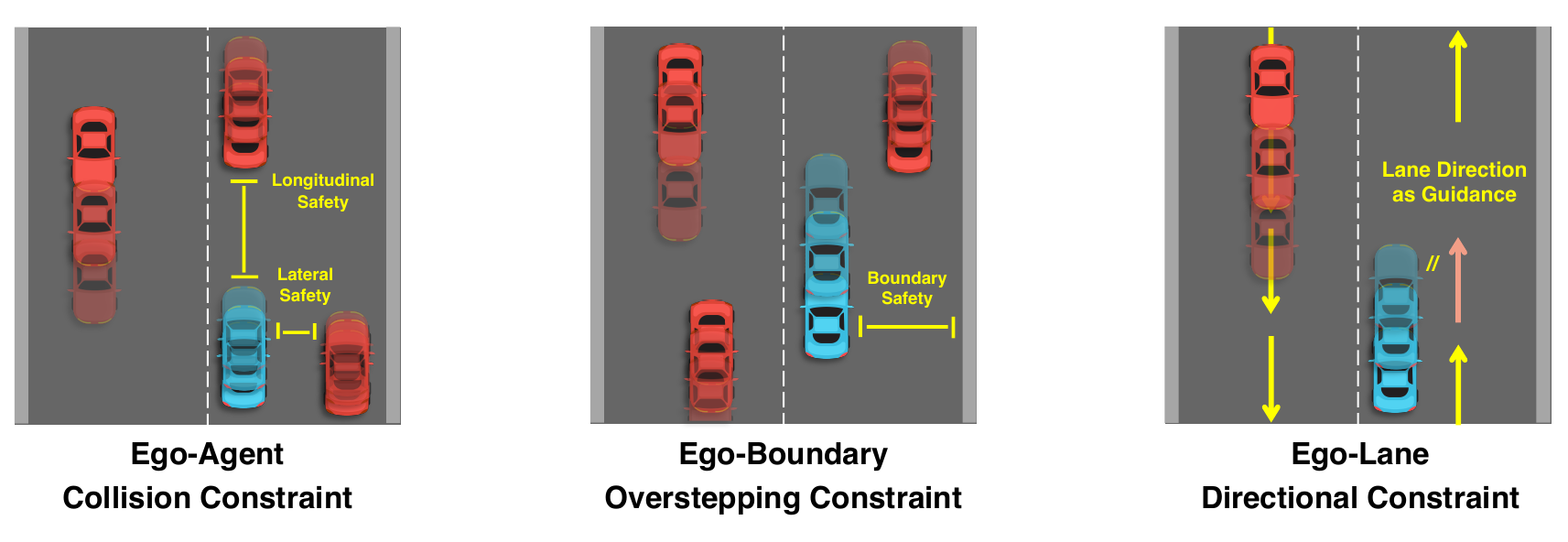} 
\caption{\textbf{Illustration of Vectorized Planning Constraints.} Ego-agent collision constraint aims to keep longitudinal safety and lateral safety between the ego vehicle and other agents. Ego-boundary overstepping constraint punishes the predictions when the planning trajectory gets too close to the lane boundary. Ego-lane directional constraint leverages the direction of the closet lane vector from the ego car (the pink lane in the right sub-figure) as prior to regularize the motion direction of planning.}
\label{fig:constraints}
\end{figure*}

\paragraph{Ego-Map Interaction.}
After interacting with agent queries, the updated ego query $Q_{\rm ego}^{'}$  further interacts with the map queries $Q_{m}$ in a similar way. The only difference is we use a different MLP ${\rm PE}_{2}$ to encode the positions of the ego vehicle and the map elements. The output ego query $Q_{\rm ego}^{''}$  contains both dynamic and static information of the driving scene. The process is formulated as:

\begin{equation}
\begin{aligned}
\label{eq:ego-map}
    &Q_{\rm ego}^{''} = {\rm TransformerDecoder}(q, \, k, \, v, \, q_{\rm pos}, \, k_{\rm pos}), \\
    &q = Q_{\rm ego}^{'}, \, k = v = Q_{m}, \\
    &q_{\rm pos} = {\rm PE}_{2}(p_{\rm ego}), \, k_{\rm pos}={\rm PE}_{2}(p_{m}).
\end{aligned}
\end{equation}

\paragraph{Planning Head.}
Because VAD performs HD-map-free planning, a high-level driving command $c$ is required for navigation. Following the common practice~\cite{hu2022stp3, hu2022uniad}, VAD uses three kinds of driving commands: \textit{turn left}, \textit{turn right} and \textit{go straight}. So the planning head takes the updated ego queries ($Q_{\rm ego}^{'}, Q_{\rm ego}^{''}$) and the current status of the ego vehicle $s_{\rm ego}$ (optional) as ego features $f_{\rm ego}$, as well as the driving command $c$ as inputs, and outputs the planning trajectory $\hat{V}_{\rm ego} \in \mathbb{R}^{T_f \times 2}$. VAD adopts a simple MLP-based planning head. The decoding process is formulated as follows:

\begin{equation}
\begin{aligned}
\label{eq:plan}
    &\hat{V}_{\rm ego} = {\rm PlanHead}({\rm ft}=f_{\rm ego}, \ {\rm cmd}=c), \\
    &f_{\rm ego} = [Q_{\rm ego}^{'}, \ Q_{\rm ego}^{''}, \ s_{\rm ego}].
\end{aligned}
\end{equation}
where $[...]$ denotes concatenation operation, ${\rm ft}$ denotes features used for decoding, and ${\rm cmd}$ denotes the navigation driving command.

\subsection{Vectorized Planning Constraint}
\label{sec:plan_loss}
Based on the learned map vector and motion vector, 
VAD regularizes the planning trajectory $\hat{V}_{\rm ego}$  with instance-level vectorized constraints during the training phase, as shown in Fig.~\ref{fig:constraints}.
\paragraph{Ego-Agent Collision Constraint.}
Ego-agent collision constraint explicitly considers the compatibility of the ego planning trajectory and the future trajectory of other vehicles, in order to improve planning safety and avoid collision. Unlike previous works~\cite{hu2022stp3, hu2022uniad} that adopt dense occupancy maps, we utilize vectorized motion trajectories which both keep great interpretability and require less computation. Specifically, we first filter out low-confidence agent predictions by a threshold $\epsilon_a$. For multi-modality motion prediction, we use the trajectory with the highest confidence score as the final prediction. We consider collision constraint as a safe boundary for the ego vehicle both laterally and longitudinally. Multiple cars may be close to each other (\eg, driving side by side) in the lateral direction,  but a longer safety distance is required in the longitudinal direction. So we adopt different agent distance thresholds $\delta_{X}$ and $\delta_{Y}$ for different directions. For each future timestamp, we find the closest agent within a certain range $\delta_{a}$ in both directions. Then for each direction $i \in \{{\rm X, Y}\}$, if the distance $d_{a}^{i}$ with the closet agent is less than the threshold $\delta_{i}$, then the loss item of this constraint $\mathcal{L}_{\rm col}^{i} = \delta_{i} - d_{a}^{i}$, otherwise it is 0. The loss for ego-agent collision constraint can be formulated as:

\begin{equation}
\begin{aligned}
\label{loss:safety}
    & \mathcal{L}_{\rm col} = \frac{1}{T_{f}} \sum_{t=1}^{T_{f}} \sum_{i} \mathcal{L}_{\rm col}^{it}, \ i \in \{{\rm X, Y}\}, \\
    & \mathcal{L}_{\rm col}^{it} = 
    \begin{cases}
    \delta_{i} - d_{a}^{it}, & {\rm if} \ d_{a}^{it} < \delta_{i} \\
    0, & {\rm if} \ d_{a}^{it} \ge \delta_{i}.
    \end{cases}  
\end{aligned}
\end{equation}

\paragraph{Ego-Boundary Overstepping Constraint.}
This constraint aims to push the planning trajectory away from the road boundary so that the trajectory can be kept in the drivable area. We first filter out low-confidence map predictions according to a threshold $\epsilon_m$. Then for each future timestamp, we calculate the distance $d_{bd}^{t}$ between the planning waypoint and its closest map boundary line. Then the loss for this constraint is formulated as:

\begin{equation}
\begin{aligned}
\label{loss:map_bd}
    & \mathcal{L}_{\rm bd} = \frac{1}{T_{f}} \sum_{t=1}^{T_{f}} \mathcal{L}_{\rm bd}^{t}, \\
    & \mathcal{L}_{\rm bd}^{t} = 
    \begin{cases}
    \delta_{bd} - d_{bd}^{t}, & {\rm if} \ d_{bd}^{t} < \delta_{bd} \\
    0, & {\rm if} \ d_{bd}^{t} \ge \delta_{bd},
    \end{cases}  
\end{aligned}
\end{equation}
 where $\delta_{bd}$ is the map boundary threshold.

\begin{table*}[]
\begin{center}
\begin{tabular}{l|cccc|cccc|cc}
\toprule
\multirow{2}{*}{Method} &
\multicolumn{4}{c|}{L2 (m) $\downarrow$} & 
\multicolumn{4}{c|}{Collision (\%) $\downarrow$} &
\multirow{2}{*}{Latency (ms)} &
\cellcolor{gray!30} \\
& 1s & 2s & 3s & \cellcolor{gray!30}Avg. & 1s & 2s & 3s & \cellcolor{gray!30}Avg. & & \cellcolor{gray!30}\multirow{-2}*{FPS} \\
\midrule
NMP$^\dagger$~\cite{zeng2019nmp} & - & - & 2.31 & \cellcolor{gray!30}- & - & - & 1.92 & \cellcolor{gray!30}- & - & \cellcolor{gray!30}- \\
SA-NMP$^\dagger$~\cite{zeng2019nmp} & - & - & 2.05 & \cellcolor{gray!30}- & - & - & 1.59 & \cellcolor{gray!30}- & - & \cellcolor{gray!30}- \\
FF$^\dagger$~\cite{hu2021ff} & 0.55 & 1.20 & 2.54 & \cellcolor{gray!30}1.43 & 0.06 & 0.17 & 1.07 & \cellcolor{gray!30}0.43 & - & \cellcolor{gray!30}- \\
EO$^\dagger$~\cite{khurana2022eo} & 0.67 & 1.36 & 2.78 & \cellcolor{gray!30}1.60 & 0.04 & 0.09 & 0.88 & \cellcolor{gray!30}0.33 & - & \cellcolor{gray!30}- \\
\midrule
ST-P3~\cite{hu2022stp3} & 1.33 & 2.11 & 2.90 & \cellcolor{gray!30}2.11 & 0.23 & 0.62 & 1.27 & \cellcolor{gray!30}0.71 & 628.3 & \cellcolor{gray!30}1.6 \\
UniAD~\cite{hu2022uniad} & 0.48 & 0.96 & 1.65 & \cellcolor{gray!30}1.03 & \textbf{0.05} & \textbf{0.17} & 0.71 & \cellcolor{gray!30}0.31 & 555.6 & \cellcolor{gray!30}1.8 \\
VAD-Tiny & 0.46 & 0.76 & 1.12 & \cellcolor{gray!30}0.78 & 0.21 & 0.35 & 0.58 & \cellcolor{gray!30}0.38 & \textbf{59.5} & \cellcolor{gray!30}\textbf{16.8} \\
VAD-Base & \textbf{0.41} & \textbf{0.70} & \textbf{1.05} & \cellcolor{gray!30}\textbf{0.72} & 0.07 & \textbf{0.17} & \textbf{0.41} & \cellcolor{gray!30}\textbf{0.22} & 224.3 & \cellcolor{gray!30}4.5 \\
\midrule
\color{gray}VAD-Tiny$^\ddagger$ & \color{gray}0.20 & \color{gray}0.38 & \color{gray}0.65 & \cellcolor{gray!30}\color{gray}0.41 & \color{gray}0.10 & \color{gray}0.12 & \color{gray}0.27 & \cellcolor{gray!30}\color{gray}0.16 & \color{gray}59.5 & \cellcolor{gray!30}\color{gray}16.8 \\
\color{gray}VAD-Base$^\ddagger$ & \color{gray}0.17 & \color{gray}0.34 & \color{gray}0.60 & \cellcolor{gray!30}\color{gray}0.37 & \color{gray}0.07 & \color{gray}0.10 & \color{gray}0.24 & \cellcolor{gray!30}\color{gray}0.14 & \color{gray}224.3 & \cellcolor{gray!30}\color{gray}4.5 \\
\bottomrule
\end{tabular}
\end{center}
\caption{\textbf{Open-loop planning performance.} VAD achieves the best end-to-end planning performance and the fastest inference speed on the nuScenes~\cite{caesar2020nuscenes} val dataset. LiDAR-based methods are denoted with $\dagger$. 
$\ddagger$ denotes taking ego status information as input. In open-loop evaluation, we deactivate ego status information for a fair comparison.
FPS of ST-P3 and VAD are measured on one NVIDIA Geforce RTX 3090 GPU. FPS of UniAD is measured on one NVIDIA Tesla A100 GPU.
}
\label{tab:sota-plan}
\end{table*}
 
\paragraph{Ego-Lane Directional Constraint.}
Ego-lane directional constraint comes from a prior that the vehicle’s motion direction should be consistent with the lane direction where the vehicle locates. The directional constraint leverages the vectorized lane direction to regularize the motion direction of our planning trajectory. Specifically, first, we filter out low-confidence map predictions according to $\epsilon_m$. Then we find the closest lane divider vector $\hat{v}_m \in \mathbb{R}^{T_f \times 2 \times 2}$ (within a certain range $\delta_{\rm dir}$) from our planning waypoint at each future timestamp. 
Finally, the loss for this constraint is the angular difference averaged over time between the lane vector and the ego vector:

\begin{equation}
\label{loss:map_dir}
    \mathcal{L}_{\rm dir} = \frac{1}{T_{f}} \sum_{t=1}^{T_{f}} {\rm F}_{\rm ang}(\hat{v}_m^{t},\ \hat{v}_{\rm ego}^{t}),
\end{equation}
in which $\hat{v}_{\rm ego} \in \mathbb{R}^{T_f \times 2 \times 2}$ is the planning ego vectors. $\hat{v}_{\rm ego}^t$ denotes the ego vector starting from the planning waypoint at the previous timestamp $t-1$ and pointing to the planning waypoint at the current timestamp $t$. ${\rm F}_{\rm ang}(v_1, v_2)$ denotes the angular difference between vector $v_1$ and vector $v_2$.

\subsection{End-to-End Learning}
\label{sec:end-to-end-learning}

\paragraph{Vectorized Scene Learning Loss.}
Vectorized scene learning includes vectorized map learning and vectorized motion prediction. For vectorized map learning, Manhattan distance is adopted to calculate the regression loss between the predicted map points and the ground truth map points. Besides, focal loss~\cite{lin2017focal} is used as the map classification loss. The overall map loss is denoted as $\mathcal{L}_{{\rm map}}$.

For vectorized motion prediction, we use $\textit{l}_1$ loss as the regression loss to predict agent attributes (location, orientation, size, \etc), and focal loss~\cite{lin2017focal} to predict agent classes. For each agent who has matched with a ground truth agent, we predict $N_k$ future trajectories and use the trajectory which has the minimum final displacement error (minFDE) as a representative prediction. Then we calculate $\textit{l}_1$ loss between this representative trajectory and the ground truth trajectory as the motion regression loss. And focal loss is adopted as the multi-modal motion classification loss. The overall motion prediction loss is denoted as $\mathcal{L}_{{\rm mot}}$.

\paragraph{Vectorized Constraint Loss.}
The vectorized constraint loss is composed of three constraints proposed in Sec.~\ref{sec:plan_loss}, \ie, ego-agent collision constraint $\mathcal{L}_{\rm col}$, ego-boundary overstepping constraint $\mathcal{L}_{\rm bd}$, and ego-lane directional constraint $\mathcal{L}_{\rm dir}$,
which regularize the planning trajectory $\hat{V}_{\rm ego}$ with vectorized scene representation.

\paragraph{Imitation Learning Loss.}
The imitation learning loss $\mathcal{L}_{\rm imi}$ is an $\textit{l}_1$ loss between the planning trajectory $\hat{V}_{\rm ego}$ and the ground truth ego trajectory $V_{\rm ego}$, aiming at guiding the planning trajectory with expert driving behavior. $\mathcal{L}_{\rm imi}$ is formulated as follows:

\begin{equation}
\label{eq:loss_plan_reg}
    \mathcal{L}_{{\rm imi}} = \frac{1}{T_{f}} \sum_{t=1}^{T_{f}} ||V^{t}_{\rm ego} - \hat{V}^{t}_{\rm ego}||_1.
\end{equation}

VAD is end-to-end trainable based on the proposed vectorized planning constraint. The overall loss for end-to-end learning is the weighted sum of vectorized scene learning loss, vectorized planning constraint loss, and imitation learning loss:

\begin{equation}
\begin{aligned}
\label{loss:overall}
    \mathcal{L} = \ & \omega_{1}\mathcal{L}_{\rm map} + \omega_{2}\mathcal{L}_{\rm mot} + \omega_{3}\mathcal{L}_{\rm col} + \\
    & \omega_{4}\mathcal{L}_{\rm bd} + \omega_{5}\mathcal{L}_{\rm dir} + \omega_{6}\mathcal{L}_{\rm imi}.
\end{aligned}
\end{equation}

\begin{table*}[]
\begin{center}
\centering
\begin{tabular}{c|ccccc|cccc|cccc}
\toprule
\multirow{2}{*}{ID} & Agent & Map & Overstep. & Dir. & Col. & \multicolumn{4}{c|}{L2 (m) $\downarrow$} &\multicolumn{4}{c}{Collision (\%) $\downarrow$}  \\
& Inter. & Inter. & Const. & Const. & Const. & 1s & 2s & 3s & \cellcolor{gray!30}Avg. & 1s & 2s & 3s & \cellcolor{gray!30}Avg. \\
\midrule
1 & \cmark & - & \cmark & \cmark & \cmark & 0.52 & 0.84 & 1.22 & \cellcolor{gray!30}0.86 & 0.12 & 0.24 & 0.51 & \cellcolor{gray!30}0.29 \\
2 & - & \cmark & \cmark & \cmark & \cmark & 0.49 & 0.80 & 1.16 & \cellcolor{gray!30}0.82 & 0.07 & 0.23 & 0.49 & \cellcolor{gray!30}0.26 \\
3 & \cmark & \cmark & - & - & - & 0.43 & 0.73 & 1.11 & \cellcolor{gray!30}0.76 & 0.10 & 0.22 & 0.52 & \cellcolor{gray!30}0.28 \\
4 & \cmark & \cmark & \cmark & - & - & 0.46 & 0.78 & 1.16 & \cellcolor{gray!30}0.80 & 0.07 & 0.18 & 0.47 & \cellcolor{gray!30}0.24 \\
5 & \cmark & \cmark & - & \cmark & - & 0.42 & 0.72 & 1.10 & \cellcolor{gray!30}0.75 & 0.10 & 0.20 & 0.45 & \cellcolor{gray!30}0.25 \\
6 & \cmark & \cmark & - & - & \cmark & 0.44 & 0.75 & 1.11 & \cellcolor{gray!30}0.77 & 0.12 & 0.23 & 0.44 & \cellcolor{gray!30}0.26 \\
7 & \cmark & \cmark & \cmark & \cmark & \cmark & 0.41 & 0.70 & 1.05 & \cellcolor{gray!30}0.72 & 0.07 & 0.17 & 0.41 & \cellcolor{gray!30}0.22 \\
\bottomrule
\end{tabular}
\end{center}
\caption{\textbf{Ablation for design choices.} "Agent Inter." and "Map Inter." indicate ego-agent and ego-map query interaction in the planning module. "OverStep. Const.", "Dir. Const.", and "Col. Const." indicate Ego-Boundary Overstepping Constraint, Ego-Lane Directional Constraint, and Ego-Agent Collision Constraint, respectively.}
\label{tab:design}
\end{table*}

\section{Experiments}
\label{sec:experiemnts}
We conduct experiments on the challenging public nuScenes~\cite{caesar2020nuscenes} dataset, which contains 1000 driving scenes, and each scene roughly lasts for 20 seconds. nuScenes provides 1.4M 3D bounding boxes of 23 categories in total. The scene images are captured by 6 cameras covering 360° FOV horizontally, and the keyframes are annotated at 2Hz. Following previous works~\cite{hu2022stp3, hu2022uniad}, Displacement Error (DE) and Collision Rate (CR) are adopted to comprehensively evaluate the planning performance. For the closed-loop setting, we adopt CARLA simulator~\cite{dosovitskiy2017carla} and the Town05~\cite{prakash2021transfuser} benchmark for simulation. Following previous works~\cite{prakash2021transfuser, hu2022stp3}, Route Completion (RC) and Driving Score (DS) are used to evaluate the planning performance.

\begin{table*}[]
\begin{center}
\centering
\resizebox{0.93\textwidth}{!}{
\begin{tabular}{cccc|cccc|cccc}
\toprule
Rasterized & Vectorized & Dir. & Overstep. & \multicolumn{4}{c|}{L2 (m) $\downarrow$} &\multicolumn{4}{c}{Collision (\%) $\downarrow$}  \\
Representation & Representation & Const. & Const. & 1s & 2s & 3s & \cellcolor{gray!30}Avg. & 1s & 2s & 3s & \cellcolor{gray!30}Avg. \\
\midrule
\cmark & - & - & - & 0.43 & 0.72 & 1.08 & \cellcolor{gray!30}0.74 & 0.22 & 0.36 & 0.58 & \cellcolor{gray!30}0.39 \\
- & \cmark & - & - & 0.44 & 0.75 & 1.11 & \cellcolor{gray!30}0.77 & 0.12 & 0.23 & 0.44 & \cellcolor{gray!30}0.26 \\
- & \cmark & \cmark & \cmark & 0.41 & 0.70 & 1.05 & \cellcolor{gray!30}0.72 & 0.07 & 0.17 & 0.41 & \cellcolor{gray!30}0.22 \\
\bottomrule
\end{tabular}
}
\end{center}
\caption{\textbf{Ablation about map representation (rasterized / vectorized).} "OverStep. Const." and "Dir. Const." indicate Ego-Boundary Overstepping Constraint and Ego-Lane Directional Constraint, respectively.}
\label{tab:raster_map}
\end{table*}

\begin{table}
\begin{center}
\renewcommand{\tabcolsep}{8.0pt}
\centering
\begin{tabular}{l|cc|cc}
\toprule
\multirow{2}{*}{Method} & \multicolumn{2}{c|}{Town05 Short} & \multicolumn{2}{c}{Town05 Long}  \\
& DS $\uparrow$ & RC $\uparrow$ & DS $\uparrow$ & RC $\uparrow$ \\
\midrule
CILRS~\cite{codevilla2019CILRS} & 7.47 & 13.40 & 3.68 & 7.19 \\
LBC~\cite{chen2020lbc} & 30.97 & 55.01 & 7.05 & 32.09 \\
\color{gray}Transfuser$^\dagger$~\cite{prakash2021transfuser} & \color{gray}54.52 & \color{gray}78.41 & \textbf{\color{gray}33.15} & \color{gray}56.36 \\
ST-P3~\cite{hu2022stp3} & 55.14 & 86.74 & 11.45 & \textbf{83.15} \\
VAD-Base & \textbf{64.29} & \textbf{87.26} & \textbf{30.31} & 75.20 \\
\bottomrule
\end{tabular}
\end{center}
\vspace{-2mm}
\caption{\textbf{Closed-loop simulation results.} VAD achieves the best closed-loop vision-only end-to-end planning performance on CARLA~\cite{dosovitskiy2017carla}. $^\dagger$: LiDAR-based method.}
\label{tab:close-loop}
\end{table}

\subsection{Implementation Details}
\label{sec:detail}
VAD uses 2-second history information and plans a 3-second future trajectory. ResNet50~\cite{he2016resnet} is adopted as the default backbone network for encoding image features. VAD performs vectorized mapping and motion prediction for a $60{\rm m} \times 30{\rm m}$ perception range longitudinally and laterally. We have two variants of VAD, which are VAD-Tiny and VAD-Base. VAD-Base is the default model for the experiments. The default number for BEV query, map query, and agent query is $200 \times 200$, $100 \times 20$, and 300, respectively. There is a total of 100 map vector queries, each containing 20 map points. The feature dimension and the default hidden size are 256. Compared with VAD-Base, VAD-Tiny has fewer BEV queries, which is $100 \times 100$. The number of BEV encoder layer and decoder layer of motion and map modules is reduced from 6 to 3, and the input image size is reduced from $1280 \times 720$ to $640 \times 360$.

\begin{table}
\begin{center}
\renewcommand{\tabcolsep}{7.0pt}
\begin{tabular}{l|cc}
\toprule
Module & Latency (ms) & Proportion  \\
\midrule
Backbone & 23.2 & 39.0\% \\
BEV Encoder & 12.3 & 20.7\% \\
Motion Module & 11.5 & 19.3\% \\
Map Module & 9.1 & 15.3\% \\
\cellcolor{gray!30}Planning Module & \cellcolor{gray!30}3.4 & \cellcolor{gray!30}5.7\% \\
\midrule
Total & 59.5 & 100.0\% \\
\bottomrule
\end{tabular}
\end{center}
\caption{\textbf{Module runtime.} The inference speed is measured for VAD-Tiny on one NVIDIA GeForce RTX 3090 GPU.}
\label{tab:com_speed}
\vspace{-2mm}
\end{table}

As for training, the confidence thresholds $\epsilon_a$ and $\epsilon_m$ are set to 0.5, the distance thresholds $\delta_{\rm a}, \delta_{\rm bd}$ and $\delta_{\rm dir}$ are 3.0m, 1.0m, and 2.0m, respectively. The agent safety threshold $\delta_{X}$ and $\delta_{Y}$ are set to 1.5m and 3.0m. We use AdamW~\cite{loshchilov2017adamw} optimizer and Cosine Annealing~\cite{loshchilov2016cosineanneal} scheduler to train VAD with weight decay 0.01 and initial learning rate $2\times{10}^{-4}$. VAD is trained for 60 epochs on 8 NVIDIA GeForce RTX 3090 GPUs with batch size 1 per GPU.

\begin{figure*}[h]
\centering
\includegraphics[width=0.98\textwidth]{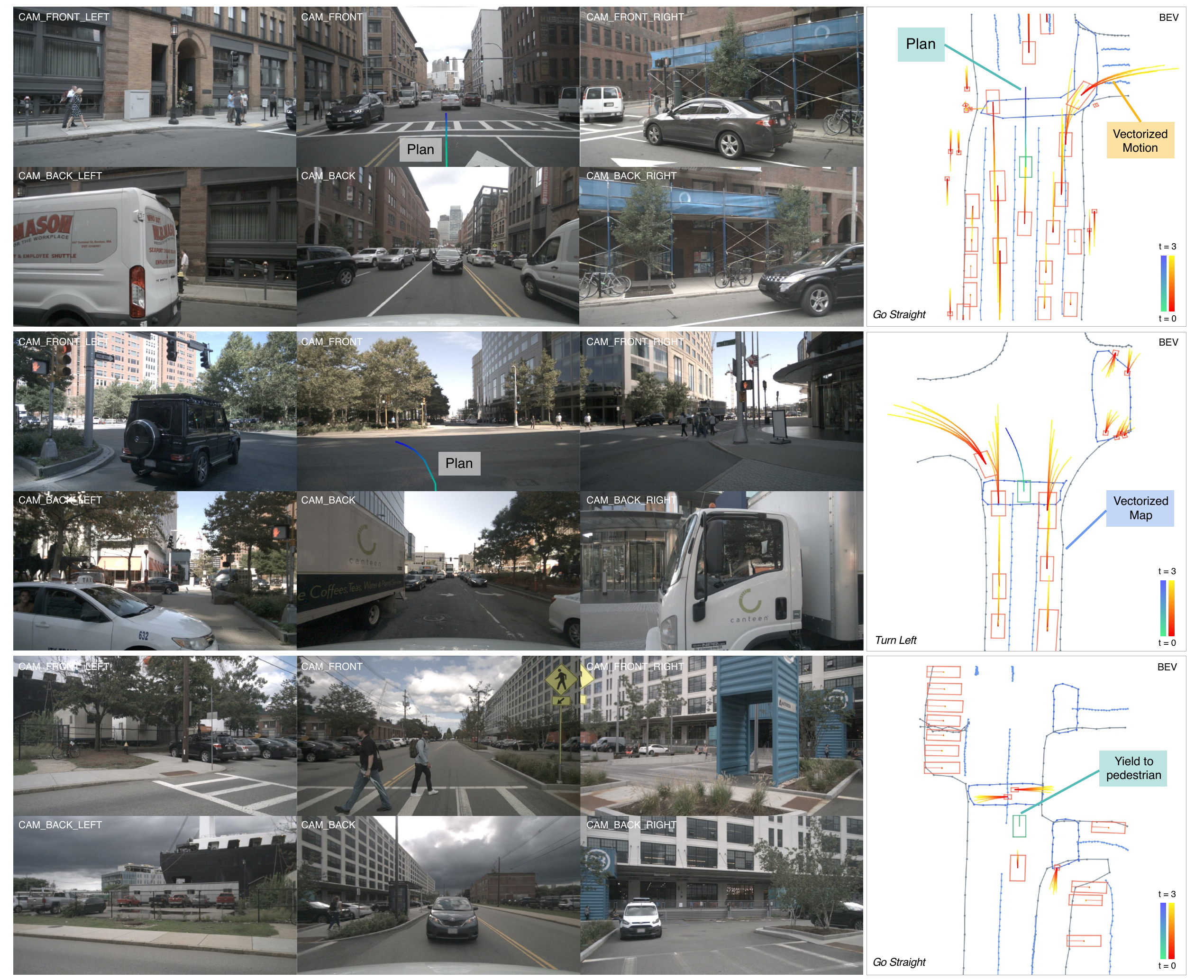} 
\caption{\textbf{Qualitative results of VAD.} VAD outputs planning results based on the vectorized scene representation in an end-to-end manner, not requiring any dense maps or hand-designed post-processing steps.}
\label{fig:vis}
\end{figure*}

VAD-Base is adopted for the closed-loop evaluation. The input image size is $640 \times 320$. Following previous works~\cite{prakash2021transfuser, hu2022stp3}, the navigation information includes a sparse goal location and a corresponding discrete navigational command. This navigation information is encoded by an MLP and sent to the planning head as one of the input features. Besides, we add a traffic light classification branch to recognize the traffic signal. Specifically, it consists of a Resnet50 network and an MLP-based classification head. The input of this branch is the cropped front-view image, corresponding to the upper middle part of the image. The image feature map is flattened and also sent to the planning head to help the model realize the traffic light information.

\subsection{Main Results}
\label{sec:main_results}
\paragraph{Open-loop planning results.}
As shown in Tab.~\ref{tab:sota-plan}, VAD shows great advantages in both performance and speed compared with the previous SOTA method~\cite{hu2022uniad}. On one hand, VAD-Tiny and VAD-Base greatly reduce the average planning displacement error by 0.25m and 0.31m. Meanwhile, VAD-Base greatly reduces the average collision rates by 29.0\%. On the other hand, because VAD does not need many auxiliary tasks (\eg, tracking and occupancy prediction) and tedious post-processing steps, it achieves the fastest inference speed based on the vectorized scene representation. VAD-Tiny runs 9.3$\times$ faster while keeping a comparable planning performance. VAD-Base achieves the best planning performance and still runs 2.5$\times$ faster.It is worth noticing that in the main results, VAD omits ego status features to avoid shortcut learning in the open-loop planning~\cite{zhai2023rethinking}, but the results of VAD using ego status features are still preserved in Tab.~\ref{tab:sota-plan} for reference.

\paragraph{Closed-loop planning results.} VAD outperforms previous SOTA vision-only end-to-end planning methods~\cite{prakash2021transfuser, hu2022stp3} on the Town05 Short benchmark. Compared to ST-P3~\cite{hu2022stp3}, VAD greatly improves DS by 9.15 and has a better RC. On the Town05 Long benchmark, VAD achieves 30.31 DS, which is close to the LiDAR-based method~\cite{prakash2021transfuser}, while significantly improving RC from 56.36 to 75.20. ST-P3~\cite{hu2022stp3} obtains better RC but has a much worse DS.

\subsection{Ablation Study}
\label{sec:ablation}
\paragraph{Effectiveness of designs.}
Tab.~\ref{tab:design} shows the effectiveness of our design choices. First, because map can provide critical guidance for planning, the planning distance error is much larger without ego-map interaction (ID 1). Second, the ego-agent interaction and ego-map interaction provide implicit scene features for the ego query so that the ego car can realize others' driving intentions and plan safely. The collision rate becomes much higher without interaction (ID 1-2). Finally, the collision rate can be reduced with any of the vectorized planning constraints (ID 4-6). When utilizing the three constraints together, VAD achieves the lowest collision rate and the best planning accuracy (ID 7).


\paragraph{Rasterized map representation.}
We show the results of a VAD variant with rasterized map representation in Tab.~\ref{tab:raster_map}. Specifically, this VAD variant utilizes map queries to perform BEV map segmentation instead of vectorized map detection, and the updated map queries are used in the planning transformer the same as VAD. As shown in Tab.~\ref{tab:raster_map}, VAD with rasterized map representation suffers from a much higher collision rate.

\paragraph{Runtime of each module.}
We evaluate the runtime of each module of VAD-Tiny, and the results are shown in Tab.~\ref{tab:com_speed}. Backbone and BEV Encoder take most of the runtime for feature extraction and transformation. Then motion module and map module take 34.6\% of the total runtime to accomplish multi-agent vectorized motion prediction and vectorized map prediction. The runtime of the planning module is only 3.4ms, thanks to the sparse vectorized representation and concise model design.

\subsection{Qualitative Results}
\label{sec:visualization}
We show three vectorized scene learning and planning results of VAD in Fig.~\ref{fig:vis}. For a better understanding of the scene, we also provide raw surrounding camera images and project the planning trajectories to the front camera image. VAD can predict multi-modality agent motions and map elements accurately, as well as plan the ego future movements reasonably according to the vectorized scene representation.

\section{Conclusion}
In this paper, we explore the fully vectorized representation of the driving scene, and how to effectively incorporate the vectorized scene information for better planning performance. The resulting end-to-end autonomous driving paradigm is termed VAD.  VAD achieves both high performance and high efficiency, which are vital for the safety and deployment of an autonomous driving system. We hope the impressive performance of VAD can reveal the potential of vectorized paradigm to the community.

VAD predicts multi-modality motion trajectories for other dynamic agents, We use the most confident prediction in our collision constraint to improve planning safety. How to utilize the multi-modality motion predictions for planning, is worthy of future discussion. Besides, how to incorporate other traffic information (\eg, lane graph, road sign, traffic light, and speed limit) into this autonomous driving system, also deserves further exploration.

{\small
\bibliographystyle{ieee_fullname}
\bibliography{egbib}
}

\end{document}